\documentclass[conference]{IEEEtran}
\usepackage{cite}
\usepackage{amsmath,amssymb,amsfonts}
\usepackage{algorithmic}
\usepackage{graphicx}
\usepackage{textcomp}
\usepackage{xcolor}
\usepackage{fancyhdr}
\usepackage[hyphens]{url}

\usepackage[para,online,flushleft]{threeparttable}
\usepackage{makecell}
\usepackage{booktabs}
\usepackage{multirow}
\usepackage{algorithm} 
\usepackage[algo2e]{algorithm2e} 

\usepackage{multirow}
\usepackage{longtable}

\def\BibTeX{{\rm B\kern-.05em{\sc i\kern-.025em b}\kern-.08em
    T\kern-.1667em\lower.7ex\hbox{E}\kern-.125emX}}

\renewcommand{\baselinestretch}{0.93}

\title{Achieving Real-Time Object Detection on  Mobile Devices with Neural Pruning Search} 
 \author{\IEEEauthorblockN{ $^1$Pu Zhao, $^2$Wei Niu, $^1$Geng Yuan, $^1$Yuxuan Cai, $^2$Bin Ren, $^1$Yanzhi Wang, $^1$Xue Lin}
\IEEEauthorblockA{
\textit{$^1$Northeastern University,  Boston, MA}\\
\textit{$^2$William \& Mary, Williamsburg, VA}\\
}}

\begin{document}
\maketitle
\pagestyle{plain}


\begin{abstract}

Object detection plays an important role in self-driving cars for security development. However, mobile systems on self-driving cars with limited computation resources lead to difficulties for object detection. To facilitate this, we propose a compiler-aware neural pruning search framework  to achieve high-speed inference on autonomous vehicles for 2D and 3D object detection. The framework automatically searches the pruning scheme and rate for each layer to find a best-suited pruning for optimizing detection accuracy and speed performance under compiler optimization. Our experiments demonstrate that for the first time, the proposed method achieves (close-to) real-time, 55ms and 99ms inference times for YOLOv4 based 2D object detection and PointPillars based 3D detection, respectively, on an off-the-shelf mobile phone with minor (or no) accuracy loss.

\end{abstract}

\section{Introduction}

As the rapid development of the autonomous vehicles,   object detection including 2D and 3D detection is one of the most important prerequisites to autonomous navigation.
It is essential to implement real-time  object detection on  autonomous vehicles due to security considerations. However, as 2D and 3D  detection are implemented with deep neural networks (DNNs) such as YOLO \cite{bochkovskiy2020yolov4} and PointPillars \cite{lang2019pointpillars}, respectively, with tremendous memory and computation requirements,  it is challenging to achieve real-time on autonomous vehicles  with limited  memory and computation resources. 

To achieve real-time object detection on edge devices with limited resources, we propose neural pruning search with compiler optimization to implement real-time 2D  detection  with YOLO \cite{bochkovskiy2020yolov4} and 3D object detection with PointPillars \cite{lang2019pointpillars} on mobile devices. We summarize our contribution as follows,

\begin{itemize}
  \item We propose to perform a novel \emph{compiler-aware neural pruning search}  with Bayesian optimization (BO), automatically determining the pruning scheme and rate (including bypass) for each individual layer. The \emph{objective} is to maximize accuracy satisfying an inference latency constraint on the target mobile device.
  \item We can achieve (close-to) real-time, 55ms and 99ms inference times for YOLOv4 based 2D detection and PointPillars based 3D detection, respectively, on an off-the-shelf mobile phone with minor (or no) accuracy loss. Our method on 2D detection notably outperforms other acceleration frameworks such as TVM \cite{chen2018tvm} and MNN \cite{Ali-MNN}, while we are the first to support 3D detection on mobile.
\end{itemize}

\section{Automatic Neural Pruning Search} 
The framework consists of two basic components: a \emph{controller} and an \emph{evaluator}. The controller first generates various \emph{pruning proposals} from the search space. Then the evaluator  evaluates their detection accuracy and speed performance. Based on the performance, the evaluator provides guidance for controller about what a satisfying pruning proposal looks like. Next the controller generates new pruning proposals with the guidance. After iterations, the controller outputs the best pruning proposal with desirable detection performance while satisfying the real-time requirement.

\vspace{-5pt}

\subsection{Controller}

The controller generates \emph{pruning proposals} from the  search space. Each pruning proposal consists of  the pruning scheme and  rate for each layer of the model, as shown in Tab. \ref{Table: search_space}.

\begin{table}[b]
\vspace{-20pt}
\caption{Search space for each DNN layer}
\small
\centering
\begin{threeparttable}
\scalebox{0.9}{
\begin{tabular}{c | c }
    \toprule
    \multirow{1}{*}{\makecell{ Pruning scheme }}     &       \{Filter \cite{zhuang2018discrimination},
      Pattern-based \cite{niu2020patdnn}, 
     Block-based \cite{dong2020rtmobile}\}  \\
\midrule
     \makecell{  Pruning   rate}   &    \{ 1$\times$, 2$\times$, 2.5$\times$, 3$\times$, 5$\times$, 7$\times$, 10$\times$, \text{skip} \}                  \\ 
      \bottomrule
\end{tabular}}
\end{threeparttable}
\begin{tablenotes}
\centering
\end{tablenotes}
\label{Table: search_space}
\vspace{-20pt}
\end{table}

\textbf{Per-layer pruning schemes:}
The controller can choose from filter (channel) pruning \cite{zhuang2018discrimination}, pattern-based pruning \cite{niu2020patdnn} and block-based pruning \cite{dong2020rtmobile} for each layer. 

\textbf{Per-layer pruning rate:}
We can choose from the list $\{ 1\times, 2\times, 2.5\times, 3\times, 5\times,  7\times, 10\times, \text{skip}\}$, where $1\times$ means the layer is not pruned, and ``skip" means bypassing this layer.

\subsubsection{Pruning Proposal Updating}
The controller generates new proposals following the replacement probability from the evaluator. 
It determines whether to replace each node in the currently best proposal according to the replacement probability. Next if  replaced, the controller chooses randomly from two  nodes with the lowest probabilities as its replacement. 
\vspace{-5pt}

\subsection{Evaluator}

The evaluator needs to evaluate  pruning proposal performance. We  define the performance measurement (reward)  as: 
{\small
\begin{equation}    
r = V - \alpha \cdot \mathrm{max}(0, t - T),
\end{equation}}%
where $V$ is the validation mean average precision (mAP) of the model, $t$ is the model inference speed or latency, which is actually measured on a mobile device with compiler optimizations. $T$ is the threshold for the latency requirement. Generally, $r$ is high when the model satisfies real-time requirement ($t<T$) with high mAP. Otherwise $r$ is small, especially when the real-time latency requirement is violated.

\subsubsection{Evaluation with BO} \label{sec: BO}

As  evaluating each proposal needs to prune and retrain the model, incurring large time cost, we use BO \cite{chen2018bayesian} to accelerate  evaluation. As shown in Algorithm \ref{alg: qb}, given a proposal pool from the controller, we first adopt BO to select a   part of proposals with potentially better performance and evaluate their  accurate   detection and speed performance, while the rest potentially weak proposals  are not evaluated. 
Thus, we reduce the number of actual evaluated proposals. 

To deal with the non-continuous and graph-like pruning proposals, we  build a Gaussian process (GP) for BO with a Weisfeiler-Lehman (WL) graph kernel \cite{shervashidze2011weisfeiler}. 
We select the proposals according to their \emph{Expected Improvement} values.

After selecting $B$ pruning proposals from the pool, we evaluate their performance using magnitude based  framework \cite{han2015learning} following their pruning proposals for each layer.

\subsubsection{Gradients Guidance} \label{sec: guidance}

To guide the  proposal updating, we  employ the derivatives of the GP predictive mean with reference to the number of nodes in the graph. Basically, positive gradients show that the node is beneficial to improve the reward, 
while negative gradients mean that the node decreases the performance and it should be replaced. 
To make the gradients more illustrative, we   transform the gradients into a probability  distribution (replacement probability) using a sigmoid transformation on the negative of the gradients  and then normalize them.
Thus, negative gradients lead to high replacement probabilities. 
To summarize, the evaluator provides the gradient guidance  including the best evaluated pruning proposal and its corresponding replacement probability obtained from its gradients. 

\begin{algorithm}[tb]
        \small
	    \caption{Evaluation with BO }\label{alg: qb}
	\begin{algorithmic}
		\STATE {\bf Input:}  Observation data $\mathcal{D}$, BO batch size $B$, BO acquisition function $\alpha(\cdot)$
		\STATE {\bfseries Output:} The best pruning proposal $g$
        \FOR{steps}
        		\STATE Generate a pool of candidate pruning proposals $ \mathcal{G}_c $;
        		\STATE Select $\{ \hat{g}^{ i} \}_{i=1}^B = \arg\max_{g \in  \mathcal{G}_c } \alpha (g \vert \mathcal{D})$;
        		\STATE Evaluate the proposal and obtain reward  $\{r^{ i}\}_{i=1}^B$ of $\{ \hat{g}^{ i} \}_{i=1}^B$;
        		\STATE Obtain the gradients guidance information;
        		\STATE $\mathcal{D}\leftarrow \mathcal{D} \cup (\{ \hat{g}^{  i} \}_{i=1}^B , \{r^{i}\}_{i=1}^B)$;
        		\STATE Update GP of BO with  $\mathcal{D}$;
    	\ENDFOR
	\end{algorithmic}
\end{algorithm}

\section{Experimental Results}


\begin{figure}[t]
    \centering
    \includegraphics[width=0.64 \columnwidth]{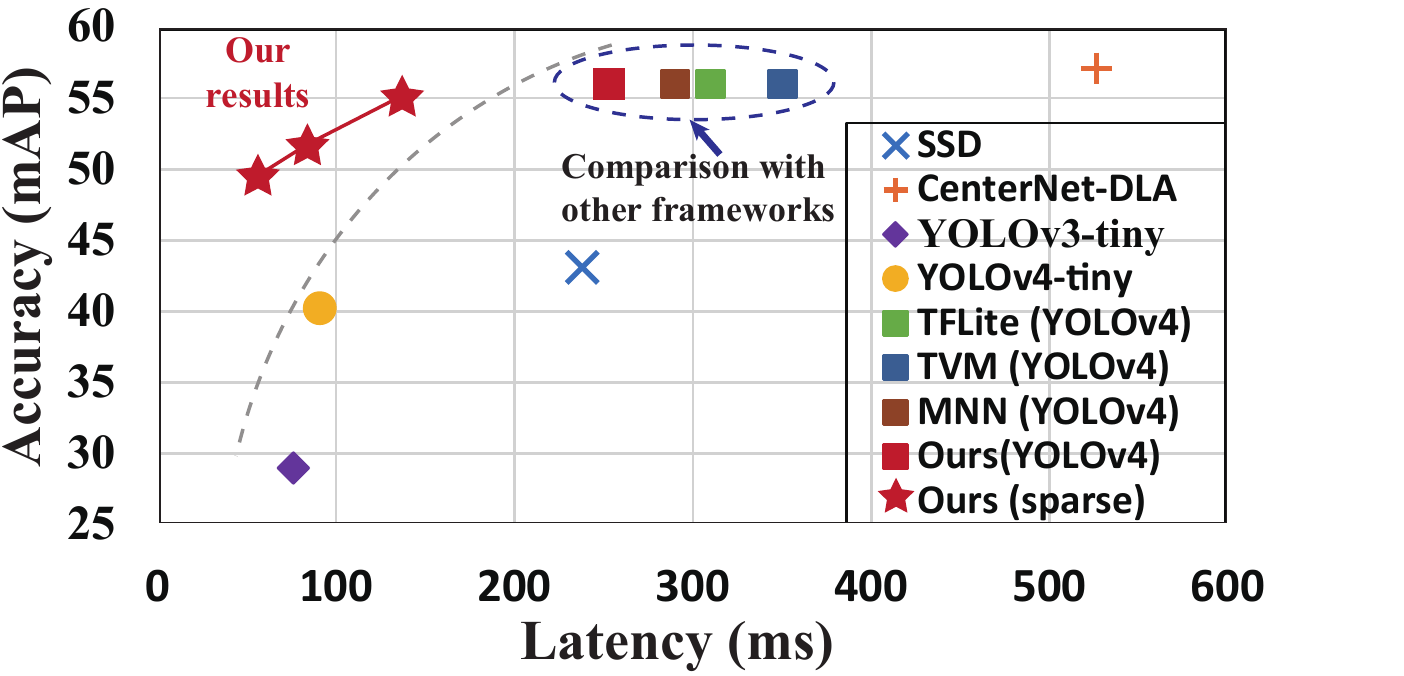}
           \vspace{-10pt}
    \caption{mAP vs. latency for various object detection approaches.}
    \label{fig: yolo_comp}
       \vspace{-15pt}
\end{figure}

For 2D object detection, we use a YOLOv4 \cite{bochkovskiy2020yolov4} model as starting point and test on COCO dataset \cite{lin2014coco}. For 3D detection, we employ the PointPillars as starting point \cite{lang2019pointpillars} and test on KITTI dataset \cite{Geiger2012CVPR}. 
All the acceleration results are tested on the mobile GPU of a Samsung Galaxy S20 smartphone.

For 2D detection, as shown in Fig. \ref{fig: yolo_comp},  on mobile GPU, our method  achieves 5.18$\times$ inference acceleration (285.7ms vs. 55.2ms) compared with the original model. 
Compared with other pruning schemes, under the same pruning rate, our method is a bit slower than filter pruning on mobile GPU but achieves much higher accuracy (49.3 vs. 25.2 in mAP). 
With slightly lower accuracy, our method is 1.79$\times$ faster than unstructured pruning.  
 We achieve faster speed compared with other acceleration frameworks such as MNN \cite{Ali-MNN}.

\begin{figure}[t]
    \centering
    \includegraphics[width=0.7 \columnwidth]{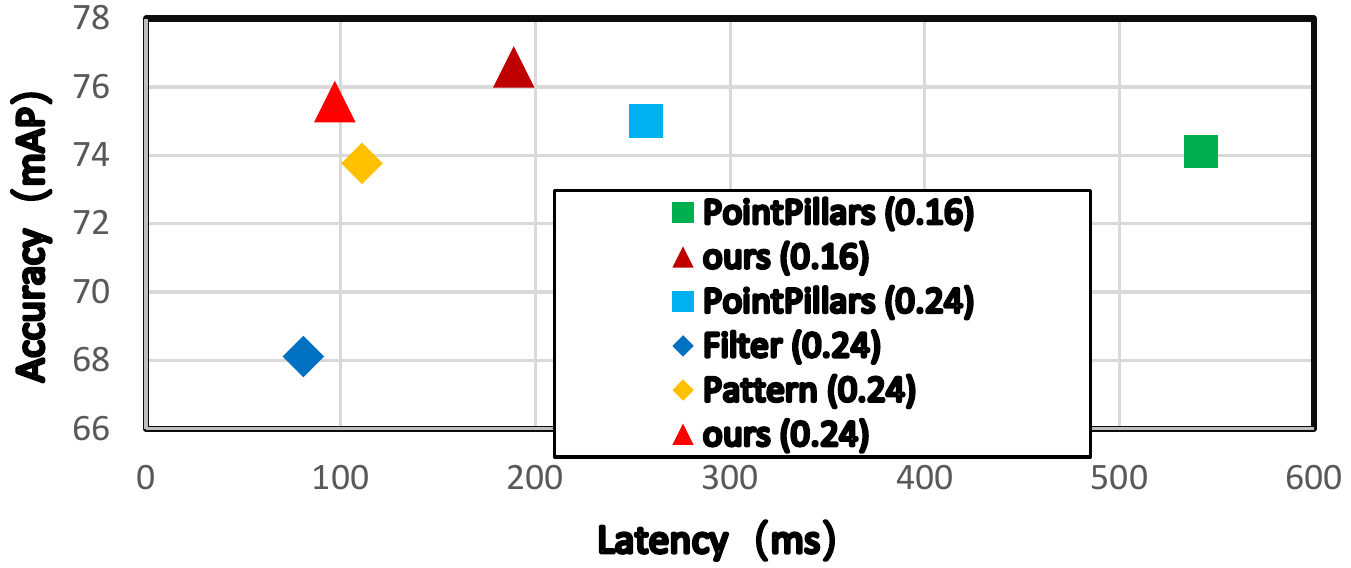}
           \vspace{-10pt}
    \caption{mAP vs. latency for various object detection approaches.}
    \label{fig: pp_comp}
       \vspace{-10pt}
\end{figure}

For 3D detection with point clouds, we start from PointPillars and test with different  different grid sizes (0.16m and 0.24m).  The real-time requirements  are set to 200ms for 0.16m gird size  and 100ms for 0.24m. 
As shown in  Tab. \ref{tab: pruning} and Fig. \ref{fig: pp_comp}, we can observe that,  for the same grid size, our method can significantly reduce the parameter count and computation, thus satisfying the real-time requirement, while achieving state-of-the-art  detection performance. 
For a grid size of 0.24m,  under the same overall pruning ratio (86\%), the proposed method can achieve the best detection performance compared with other methods with the same pruning scheme for each layer, demonstrating the advantages of using flexible pruning scheme for each layer. 
Besides, with compiler optimization, filter pruning is the fastest but suffers from obvious detection performance degradation. 
The proposed method can process one LiDAR image within  99ms with the highest precision,  achieving (close-to) real-time inference on mobile.  

\begin{table}[tb]    
\caption{Comparison of various   pruning methods for PointPillars}
	    \vspace{-15pt}
\begin{center}
\begin{threeparttable}
\scalebox{0.82}{
\begin{tabular}{c|c|c|c|c|c|c}
\toprule
\multirow{2}{*}{ \makecell{Methods \\ (grid size)}}    &
\multirow{2}{*}{\makecell{Para. \\ \#}}  & 
\multirow{2}{*}{\makecell{ Comp.  \# \\   (MACs)}}  & 
\multirow{2}{*}{\makecell{ Speed \\  (ms)}}  & \multicolumn{3}{c}{Car 3D detection} \\
\cline{5-7}
            &  &   &  & Easy      & Moderate      & Hard        \\
\midrule
             PointPillars (0.16)  & 5.8M   & 60G & 542 & 84.99 & 74.11 & 69.53  \\
                     {Ours}  (0.16)  &   1.1M & 10.7G & 189 & \textbf{85.50} & \textbf{76.58}&\textbf{70.23}  \\
    \midrule                 
          PointPillars (0.24)  & 5.8M & 28G & 257  & 84.05  & 74.99  & 68.30 \\
           Filter \cite{he2019filter}  (0.24) & 0.8M &   4.0G & 81 & 81.54 &  68.10 &  65.90  \\
           Pattern \cite{ma2019pconv}  (0.24)  & 0.8M  &  3.9G & 111 &   80.97 &  73.77 & 68.05         \\
           {Ours}  (0.24)  &  0.8M   & 3.9G  & 99 &  \textbf{85.08} & \textbf{75.19} & \textbf{68.10}   \\
\bottomrule
\end{tabular}}
\end{threeparttable}
\end{center}
\label{tab: pruning}
	    \vspace{-20pt}
\end{table}

\section{Acknowledgements}
This project is partly supported by National Science Foundation (NSF) under grants CNS-1932351, CNS-1909172, and CMMI-2013067, Army Research Office (ARO) Young Investigator Program 76598CSYIP, a grant from Semiconductor Research Corporation (SRC), and Jeffress Trust Awards in Interdisciplinary Research. Any opinions, findings, and conclusions or recommendations  in this material are those of the authors and do not necessarily reflect the views of NSF, ARO, SRC, or Thomas F. and Kate Miller Jeffress Memorial Trust. 


\bibliographystyle{IEEEtranS}
\bibliography{refs}

\end{document}